\documentclass[twoside]{article}

\usepackage[accepted]{aistats2025}
\usepackage{amsmath,amsfonts,bm}

\def\eqref#1{equation~\ref{#1}}

\def\1{\bm{1}}

\def\rv{{\textnormal{v}}}

\def\rvb{{\mathbf{b}}}
\def\rvc{{\mathbf{c}}}

\def\rvg{{\mathbf{g}}}

\def\rvv{{\mathbf{v}}}

\def\rvx{{\mathbf{x}}}
\def\rvy{{\mathbf{y}}}

\DeclareMathAlphabet{\mathsfit}{\encodingdefault}{\sfdefault}{m}{sl}
\SetMathAlphabet{\mathsfit}{bold}{\encodingdefault}{\sfdefault}{bx}{n}

\usepackage{graphics}
\usepackage[utf8]{inputenc} 
\usepackage[T1]{fontenc}    
\usepackage{hyperref}       
\usepackage{url}            
\usepackage{booktabs}       
\usepackage{amsfonts}       
\usepackage{nicefrac}       
\usepackage{microtype}      
\usepackage[dvipsnames]{xcolor}
\usepackage[utf8]{inputenc}
\usepackage{graphicx}
\usepackage{amsmath}
\usepackage{booktabs}
\usepackage{arydshln}
\usepackage{multirow}
\usepackage{wrapfig, lipsum, booktabs}
\usepackage{amsfonts}
\usepackage{algorithm}
\usepackage{algorithmic}
\usepackage{enumerate}
\usepackage{enumitem}
\usepackage{tabularx}
\usepackage{balance}
\usepackage{pgfplots}
\usetikzlibrary{pgfplots.groupplots}
\pgfplotsset{compat=1.3}
\usepackage{tikz}
\usetikzlibrary{patterns}
\usepackage{pgf-pie}
\usepackage{colortbl}
\usepackage{array}   
\usepackage{fdsymbol}
\usepackage{diagbox}
\usepackage{subfig}
\usepackage{fancybox}
\usepackage{underscore}
\usepackage{mathrsfs}
\usetikzlibrary{pgfplots.groupplots}
\pgfplotsset{compat=1.3}
\usepackage{tikz}
\usetikzlibrary{patterns}
\usepackage{pgf-pie}
\usepackage{colortbl}
\usepackage{array}   
\usepackage{fdsymbol}
\usepackage{subcaption}
\usepackage{diagbox}
\usepackage{tablefootnote}
\newlength{\offsetpage}
\definecolor{battleshipgrey}{rgb}{0.3, 0.3, 0.3}
\definecolor{brilliantrose}{rgb}{1.0, 0.33, 0.64}
\definecolor{americanrose}{rgb}{1.0, 0.01, 0.24}
\definecolor{jweigreen}{rgb}{0,0.45,0.24}
\definecolor{bluegray}{rgb}{0.1, 0.1, 0.4}
\definecolor{ao(english)}{rgb}{0.0, 0.5, 0.0}
\definecolor{blanchedalmond}{rgb}{1.0, 0.92, 0.8}
\definecolor{atomictangerine}{rgb}{1.0, 0.6, 0.4}
\definecolor{chocolate(web)}{rgb}{0.82, 0.41, 0.12}
\definecolor{bananayellow}{rgb}{1.0, 0.88, 0.21}
\definecolor{goldenbrown}{rgb}{0.6, 0.4, 0.08}
\definecolor{aliceblue}{rgb}{0.94, 0.97, 1.0}
\definecolor{beige}{rgb}{0.96, 0.96, 0.86}
\definecolor{babyblue}{rgb}{0.54, 0.81, 0.94}
\definecolor{camel}{rgb}{0.76, 0.6, 0.42}
\definecolor{cinnamon}{rgb}{0.82, 0.41, 0.12}
\definecolor{applegreen}{rgb}{0.56, 0.8, 0.25}

\usepackage{pgfplots}
\usetikzlibrary{pgfplots.groupplots}
\pgfplotsset{compat=1.3}
\usepackage{pifont}
\usepackage{microtype}

\usepackage{subcaption}

\usepackage{amsmath}

\usepackage{natbib}

\newcommand{\colorcitep}[1]{\textcolor{Blue}{\citep{#1}}}
\newcommand{\colorcitet}[1]{\textcolor{Blue}{\citet{#1}}}
\newcommand{\colorref}[1]{\textcolor{red}{\ref{#1}}}

\begin{document}

%

%

\twocolumn[

\aistatstitle{RetroDiff: Retrosynthesis as Multi-stage Distribution Interpolation}

\aistatsauthor{Yiming Wang$^{1,*}$, Yuxuan Song$^2$, Yiqun Wang$^3$, Minkai Xu$^4$, Rui Wang$^1$, Hao Zhou$^{2,\dagger}$, Wei-Ying Ma$^2$ }

\aistatsaddress{$^1$Department of Computer Science and Engineering, Shanghai Jiao Tong University\\
  $^2$Institute for AI Industry Research (AIR), Tsinghua University\\
  $^3$ByteDance Research ~~$^4$Department of Computer Science, Stanford University\\\\
   {\small $^*$Work done during Yiming's internship at AIR, Tsinghua University. $^\dagger$Corresponding Author}}

]

\begin{abstract}
Retrosynthesis poses a key challenge in biopharmaceuticals, aiding chemists in finding appropriate reactant molecules for given product molecules. With reactants and products represented as 2D graphs, retrosynthesis constitutes a conditional graph-to-graph (G2G) generative task.
Inspired by advancements in discrete diffusion models for graph generation, we aim to design a diffusion-based method to address this problem. 
However, integrating a diffusion-based G2G framework while retaining essential chemical reaction template information presents a notable challenge.
Our key innovation involves a multi-stage diffusion process. We decompose the retrosynthesis procedure to first sample external groups from the dummy distribution given products, then generate external bonds to connect products and generated groups.
Interestingly, this generation process mirrors the reverse of the widely adapted semi-template retrosynthesis workflow, \emph{i.e.} from reaction center identification to synthon completion.
Based on these designs, we introduce Retrosynthesis Diffusion (RetroDiff), a novel diffusion-based method for the retrosynthesis task.
Experimental results demonstrate that RetroDiff surpasses all semi-template methods in accuracy, and outperforms template-based and template-free methods in large-scale scenarios and molecular validity, respectively.
Code: \url{https://github.com/Alsace08/RetroDiff}.
\end{abstract}

\section{Introduction}
\label{sec:intro}

\begin{figure*}[htbp]
\begin{center}
\vspace{-0.1in}
    \begin{minipage}[b]{1.8\columnwidth}
        \centering
        \subfloat[{\bf RetroDiff Pipeline}: Macro Denoising Process]{\includegraphics[width=1.05\columnwidth,height=5.0cm]{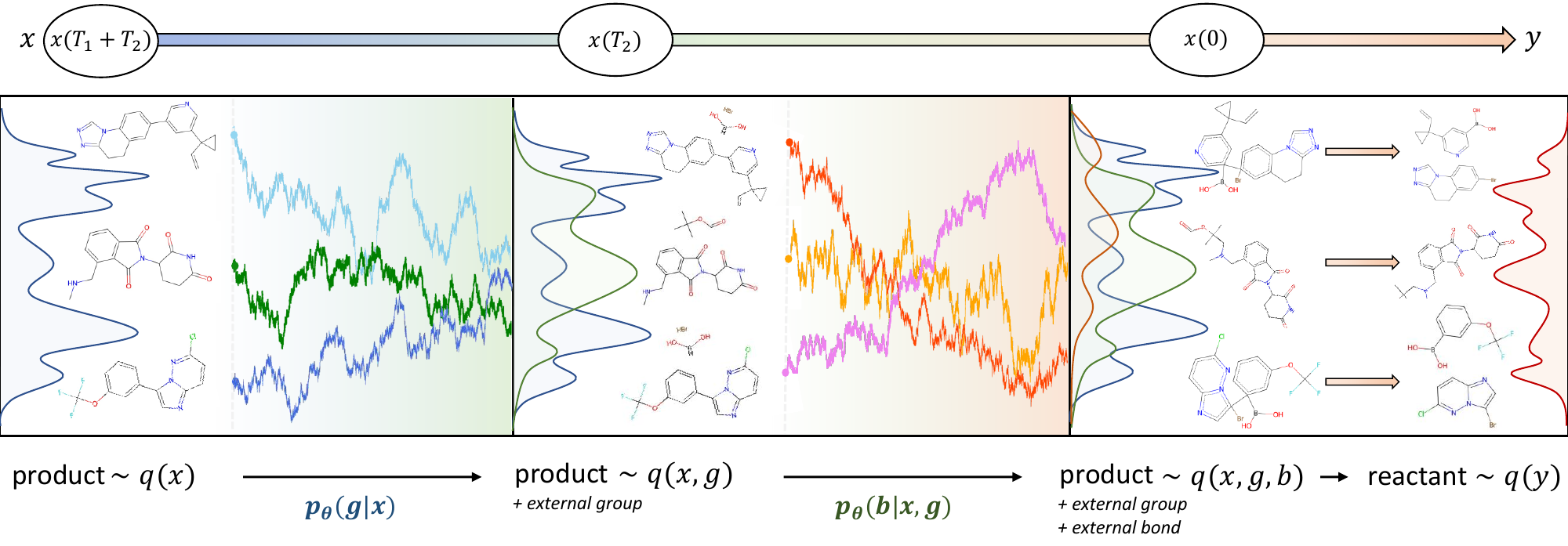}}
    \end{minipage}
    \begin{minipage}[b]{1.8\columnwidth}
        \centering
        \subfloat[{\bf RetroDiff Example}: Micro Noise-applying and Denoising Process]{\includegraphics[width=1.05\columnwidth,height=5.0cm]{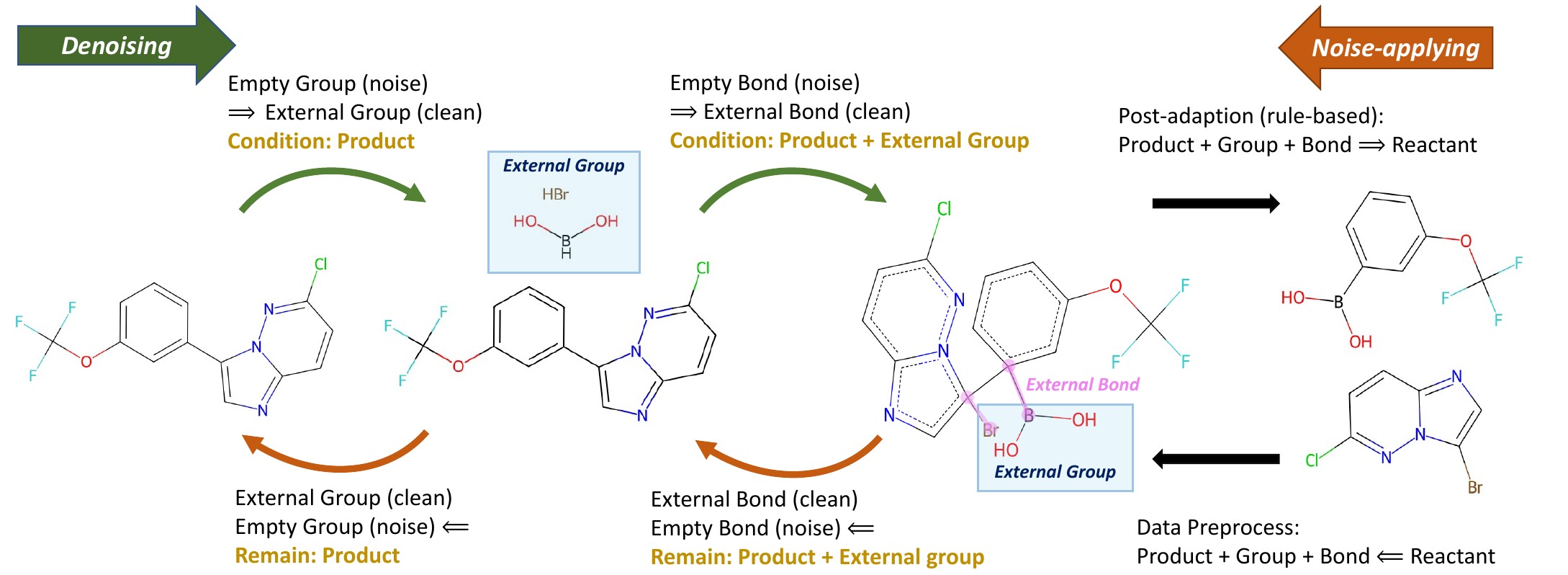}}
    \end{minipage}
    \vspace{0.0in}
    \caption{The pipeline and example of our RetroDiff model.}
\vspace{-0.2in}
\label{img:pipeline}
\end{center}
\end{figure*}

Retrosynthesis \colorcitep{corey1991logic} is important in organic synthesis, which helps chemists find legitimate reactant molecules given product molecule, thus providing efficient and stable drug discovery and compound preparation methods for the biopharmaceutical field.
Since the first computer-aided method was proposed \colorcitep{corey1969computer}, huge efforts have been devoted to exploring analytical computational methods for retrosynthesis, and research for data-driven methods has reached its peak in recent years.

Retrosynthesis methods can be broadly categorized into three groups.
\textit{Template-based} methods retrieve the best match reaction template for a target molecule from large-scale chemical databases \colorcitep{schneider2016s,chen2021deep}. 
Though with appealing performance, the scalability of template-based methods is indeed limited by the template database size \colorcitep{segler2017modelling,segler2018planning};
\textit{Template-free} methods generate the reactants given corresponding products directly without any chemical prior \colorcitep{zheng2019predicting,seo2021gta,tu2022permutation}, but limited chemical reaction diversity and interpretability hinder the potential of them in practical applications \colorcitep{chen2019learning,he2018sequence,jiang2018sequence,roberts2020decoding}.

Fortunately, \textit{semi-template} methods could be another alternative for building retrosynthesis models. 
Combining the strengths of both template-based and -free methods, semi-template methods introduce the chemical prior into models by employing a two-stage process including ``reaction center prediction'' and ``synthon completion''.
This makes semi-template methods more scalable than template-based ones and more interpretable than template-free ones, which has drawn increasing interest of late \colorcitep{yan2020retroxpert,shi2020graph,wang2021retroprime}.
In this paper, we aim to develop a more effective semi-template method.

The non-autoregressive generative model diffusion is particularly well-suited for capturing the complex structure of graph data, along with its robust capability for probabilistic modeling. 
However, the intractable reactant and product distributions impede a naive adoption of diffusion models to smoothly interpolate between these chemical spaces.
Moreover, current reaction templates overly constrain the intrinsic data structure and necessitate artificial modifications to the molecular structure of groups and bonds, making it difficult to provide the explicit product prior for the diffusion modeling.
To address this issue, we redefine the reaction template by separating the external group generation from the external bond generation. This revised approach aligns with the concept of retrosynthesis, wherein the task is to transform distributions with minimal constraints: {\it given a product molecule, we generate a dummy distribution that transitions to distributions of external groups and bonds, then we splice these to form the reactant distribution.}
With such a template setup, we cleverly assign the intractable product distribution to learning conditions rather than goals via chemical prior.

Building on this template, we introduce \textbf{RetroDiff} --- a \textbf{Retro}synthesis \textbf{Diff}usion model that works in discrete conditions, as illustrated in Figure \colorref{img:pipeline}.
The model first generates molecular structures through a two-stage denoising process: Initially, it begins with a prior distribution, proceeding first to create the external groups, which are parts that attach to the product molecule (Sec.\colorref{sec:group});
Once these groups are formed, the model then constructs the bonds that connect these external groups to the product (Sec.\colorref{sec:bond}). 
Finally, We manually remove some product bonds based on the reaction sites identified by the generated external bonds, thereby ensuring the resulting reactant is chemically valid (Sec.\colorref{sec:post-adaption}). 
RetroDiff innovatively flips the script on the conventional semi-template methods: In our method, the high-uncertainty variables (groups) are first generated, this significantly minimizes the error buildup of generating low entropy variables (bonds).

We conduct extensive experiments (Sec.\colorref{sec:main-results}) on the USPTO-50k \colorcitep{schneider2016s} and USPTO-full \colorcitep{usptofull} dataset, and empirical results show that our model achieves {\bf state-of-the-art top-$k$ performance} compared with other competitive {\it semi-template} methods.
When compared with {\it template-free} methods, our method is competitive but exhibits a {\bf higher validity} advantage, which ensures our stronger availability and security in real scenarios.
When compared with {\it template-based} methods, our method has a slight performance disadvantage on USPTO-50k. However, when the application scene is further scaled to include a larger chemical space, our method's performance far exceeds other template-based methods, which is verified on the large-scale USPTO-full, indicating that we achieve {\bf greater scalability}.

\vspace{-0.02in}
\section{RetroDiff: Retrosynthesis Diffusion}
\vspace{-0.02in}

We begin by defining the task of retrosynthesis prediction.
Consider a chemical reaction expressed as $\{\bm{G}_R^i\}_{i=1}^{|R|} \to \{\bm{G}_P^i\}_{i=1}^{|P|}$, where $\bm{G}_R$ represents the set of reactant molecular graphs, $\bm{G}_P$ represents the set of product molecular graphs, and $|R|$ and $|P|$ indicate the respective counts of reactants and products in a given reaction. Typically, we assume $|P| = 1$, which aligns with the conventions of benchmark datasets.
The key problem in the retrosynthesis task is to invert the chemical reaction; namely deduce the reactant set $\{\bm{G}_R^i\}_{i=1}^{|R|}$ when presented with a sole product $\{\bm{G}_P\}$.
In general, the assorted connected sub-graphs comprising the reactants can be amalgamated into a single disjoint graph $\{\bm{G}_R\}$. 
Thus, the retrosynthesis prediction simplifies to the transformation $\{\bm{G}_P\} \to \{\bm{G}_R\}$.

Existing semi-template methods typically first identify the reaction center in the given product and then complete the synthons at the fractured site. 
However, such a template setup is infeasible for designing the appropriate generative diffusion process.
To address this, we redefine the task template with the following preliminary notations:
$\rvx \sim P_\mathcal{X}$ denotes the variable of product graphs and the corresponding distribution, $\rvy \sim P_\mathcal{Y}$ as the reactant variable, $\rvg \sim P_\mathcal{G}$ as the external group, and $\rvb \sim P_\mathcal{B}$ as the external bond. 
We elaborate on the redefined template in the following stages:

\begin{itemize}[leftmargin=15px]
\vspace{-0.1in}
\item \textbf{Stage 1: External Group Generation.} The process commences with the generation of external group $\rvg$ that will attach to the product $\rvx$. Namely sampling from such distribution 
$P_{\mathcal{G}}(\rvg|\rvx;\theta)$ which is parameterized by the neural network $\theta$. 
\vspace{-0.05in}
\item \textbf{Stage 2: External Bond Generation.} Next, the process involves the generation of the external bond $\rvb$, which will link the product $\rvx$ with the newly formed external group $\rvg$. Here, we focus on modeling the distribution $P_{\mathcal{B}}(\rvb|\rvg, \rvx;\theta)$. 
\vspace{-0.05in}
\item \textbf{Stage 3: Post-Adaptation (Rule-Based).} The concluding phase involves a manual adjustment, breaking the reaction center in the product in line with valence rules to yield the final reactant $\rvy$. This transformation is depicted as $P_{\mathcal{Y}}(\rvy|\rvb,\rvg,\rvx)$ which is a predetermined rule-based mapping.
\end{itemize}

\vspace{-0.05in}
Building on this framework, we introduce RetroDiff, which integrates the above stages into a unified diffusion model. 
This serial procedure essentially implies an autoregressive decomposition of the probabilistic model, aimed at approximating  the conditional distribution:

\begin{equation}
    \begin{aligned}
        &P_{\text{model}}(\rvy|\rvx; \theta) \\
        &= 
        \int P_{\mathcal{G}}(\rvg|\rvx;\theta)P_{\mathcal{B}}(\rvb|\rvg, \rvx;\theta)P_{\mathcal{Y}}(\rvy|\rvb,\rvg,\rvx)~\mathrm{d}\rvb~\mathrm{d}\rvg ,
    \end{aligned}
\label{eq:stage}
\end{equation}

which essentially represents the transformation between distributions of product and reactant.

\begin{figure*}[t]
\vspace{-0.1in}
  \centering
  \includegraphics[width=1\textwidth]{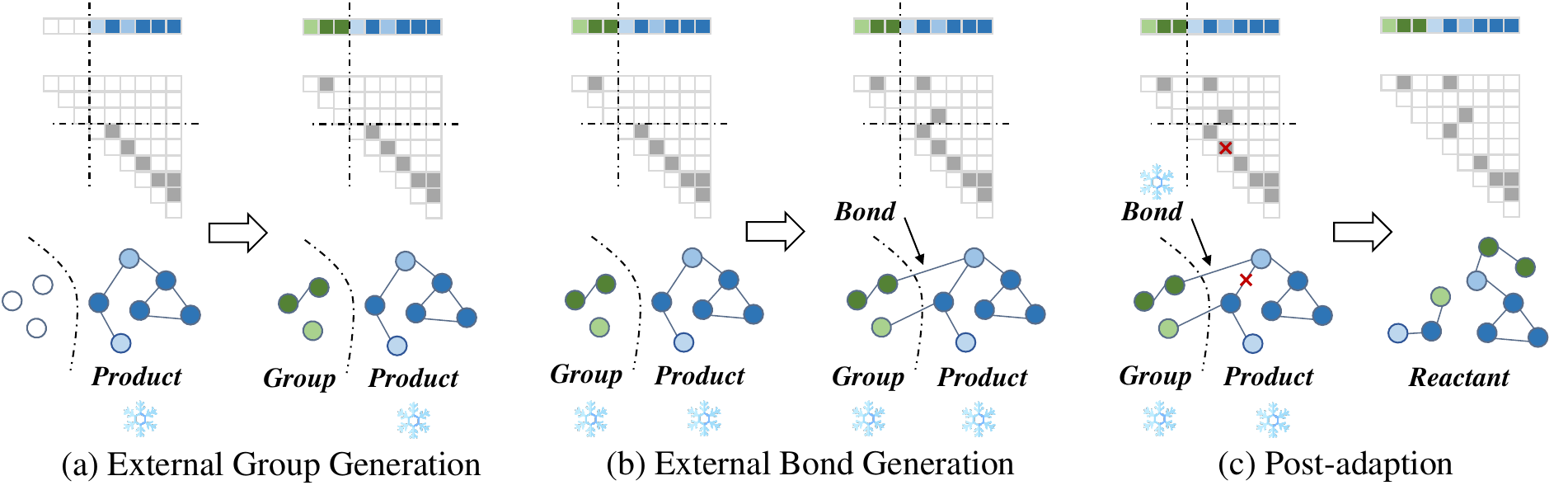}  
  \vspace{-0.25in}
  \caption{The generation overview of the distribution transformation upon our template. The top row indicates changes in the atom types in the graph, the middle row indicates changes in the adjacency matrix of the graph, and the bottom row indicates overall changes in the graph structure. Specifically, the hollow circle denotes a dummy atom category we set for this task, and the colored circles denote real atom categories. The Line between circles means there exists one bond between the two atoms.}
  \label{fig:pipeline-detail}
\vspace{-0.15in}
\end{figure*}

\vspace{-0.05in}
\subsection{RetroDiff Pipeline}
\vspace{-0.05in}

In this section, we introduce the whole pipeline of the proposed RetroDiff which includes the detailed implementations of $P_{\mathcal{G}}(\rvg|\rvx;\theta)$, $P_{\mathcal{B}}(\rvb|\rvg, \rvx;\theta)$ and $P_{\mathcal{Y}}(\rvy|\rvb,\rvg,\rvx)$, as presented in the Eq.\colorref{eq:stage}.

We utilize the diffusion process to model all the conditional distributions.
For completeness, we elaborate on the details for parameterizing the conditional distribution with a diffusion process. 
We take $P_{\mathcal{G}}(\rvg|\rvx;\theta)$ as an example. Under the context of diffusion models, the dimensions of the input and output variables should be aligned. Hence,, we append a dummy noisy variable $\rv_1$, which makes the input $(\rv_1,\rvx)$; correspondingly, the output is $(\rvg,\rvx)$. Note that here we have $\dim(\rv_1)= \dim(\rvg)$. Similarly, for $P_{\mathcal{Y}}(\rvy|\rvb,\rvg,\rvx)$, the input is $(\rv_2,\rvg, \rvx)$ while the output is as $(\rvb,\rvg,\rvx)$. 
For the training objective, we only calculate the objective on the variables concerned, $\rvg$ in $P_{\mathcal{G}}(\rvg|\rvx;\theta)$ and $\rvb$ in $P_{\mathcal{B}}(\rvb|\rvg, \rvx;\theta)$. Strictly, our model implies a transformation in the joint space as: 
\begin{equation}
    \mathcal{X} \times {\mathcal{V}_1} \times {\mathcal{V}_2} \rightarrow \mathcal{X} \times \mathcal{G} \times {\mathcal{V}_2} \rightarrow \mathcal{X} \times \mathcal{G} \times \mathcal{B} \rightarrow \mathcal{Y},
\end{equation}
Details of the generation pipeline can be found in Figure \colorref{fig:pipeline-detail} (Denoising Process).
To simplify the representation, we denote the condition at each stage as $\rvc$.

\vspace{-0.05in}
\subsubsection{External Group Generation}
\label{sec:group}
\vspace{-0.05in}
The goal of this stage is to interpolate the distribution $P_{\mathcal{V}_1}$ to $P_\mathcal{G}$ conditioned on $c$.
In this stage, condition $\rvc$ is the product $\rvx \sim P_\mathcal{X}$. 
This is a \textbf{graph-to-graph} generative process, we define $\rvv \sim P_{\mathcal{V}_1}$ as a dummy noisy graph and $\rvg \sim P_\mathcal{G}$ as a true external group.

\vspace{-0.16in}
\paragraph{Noise-applying.} 
In the noise-applying process, we interpolate the distribution $P_\mathcal{G}$ to $P_{\mathcal{V}_1}$.
With a slight abuse of notation, we splice external group $\rvg$ and product $\rvx$ into one unconnected graph $\bm{G} = (\bm{X}, \bm{E})$ with $n$ atoms and $m$ bonds, each atom and bond have $a$ and $b$ categories, respectively, so they can be represented by one-hot attributes that $\bm{X} \in \mathbb{R}^{n \times a}$ and $\bm{E} \in \mathbb{R}^{n \times n \times b}$.
For graph $\bm{G}$, each atom and bond are diffused independently \colorcitep{vignac2022digress}, which means the state transition each time acts on the single atom $x_i \in \bm{X}$ and bond $e_i \in \bm{E}$.

We follow \colorcitep{austin2021structured} to define the Markov matrix $\bm{Q}_t$ to conduct probability transitions of states at each time $t$ in the discrete space.
For graph $\bm{G}$, we apply noise to atoms via $[\bm{Q}_t^X]_{ss'} = q(x_t = s' | x_{t-1} = s)$ and bonds via $[\bm{Q}_t^E]_{ss'} = q(e_t = s' | e_{t-1} = s)$, where $s$ and $s'$ represent the atom/bond state at time $t-1$ and $t$, respectively.
Due to the graph independence, the noise-applying process for graph $\bm{G}$ can be defined as:
\begin{equation}
    \begin{aligned}
        q(\bm{G}_t | \bm{G}_{t-1}) &= (\bm{X}_{t-1} \bm{Q}_t^X, \bm{E}_{t-1} \bm{Q}_t^E) \\
        ~&\Longrightarrow~ q(\bm{G}_t | \bm{G}_{0}) = (\bm{X}_{0} \bar{\bm{Q}}_t^X, \bm{E}_{0} \bar{\bm{Q}}_t^E),
    \end{aligned}
\end{equation}
where $\bm{G}_0$ is the graph of ground truth, $\bar{\bm{Q}}_t^X = \prod_{i=1}^t \bm{Q}_t^X$ and $\bar{\bm{Q}}_t^E = \prod_{i=1}^t \bm{Q}_t^E$.
Finally, we sample the probability distribution $q(\bm{G}_t | \bm{G}_{0})$ to obtain the noisy graph $\bm{G}_t$.

\vspace{-0.1in}
\paragraph{Denoising.} 
In the denoising process, given a noisy graph $\bm{G}_t$ and condition $\rvc$, 
we need to iterate the denoising process $p_{\theta}(\bm{G}_{t-1} | \bm{G}_{t}, \rvc)$ by a trainable network $p_\theta$ at each time $t$.
We model the distribution as the product over nodes and edges and marginalize each item over the network predictions:
\begin{equation}
\label{eq:denoise}
    \begin{aligned}
        p_{\theta}(\bm{G}_{t-1} | \bm{G}_t, \rvc) =  \prod_{x \in \bm{X}_{t-1}} p_{\theta}(x | \bm{G}_t, \rvc)
        \prod_{e \in \bm{E}_{t-1}} p_{\theta}(e | \bm{G}_t, \rvc),
    \end{aligned}
\end{equation}
where
\begin{equation}
\label{eq:parameter}
    \begin{aligned}
        &p_{\theta}(x | \bm{G}_t, \rvc) =  \sum_{x_0 \in \bm{X}_0} q(x | x_t, x_0, \rvc) p_{\theta}(x_0|\bm{G}_t, c)
        , \\
        & p_{\theta}(e | \bm{G}_t, \rvc) = \sum_{e_0 \in \bm{E}_0} q(e | e_t, e_0, \rvc) p_{\theta}(e_0|\bm{G}_t, c)
        . \\
    \end{aligned}
\end{equation}
Next, we derive $q(\bm{G}_{t-1} | \bm{G}_t, \bm{G}_0, \rvc)$ with the Bayes theorem and transform it into forms of node and edge to complete the calculations in Eq.\colorref{eq:denoise} \colorcitep{vignac2022digress}. For node $X$, we have:
\begin{equation}
\label{eq:bayes1}
    \begin{aligned}
        &q(\bm{X}_{t-1} | \bm{X}_t, \bm{X}_0, \rvc)
        = \frac{q(\bm{X}_t | \bm{X}_{t-1}, \bm{X}_0, \rvc) q(\bm{X}_{t-1} |\bm{X}_0, c)}{q(\bm{X}_t | \bm{X}_0, \rvc)}
        \\
        & 
        = \frac{\bm{X}_t [\bm{Q}_{t}^X]^\top \odot \bm{X}_0 \bar{\bm{Q}}_{t-1}^X}{\bm{X}_0 \bar{\bm{Q}}_{t}^X [\bm{X}_{t}]^\top} 
        \propto \bm{X}_t [\bm{Q}_{t}^X]^\top \odot \bm{X}_0 \bar{\bm{Q}}_{t-1}^X.
    \end{aligned}
\end{equation}
Similarly, $q(\bm{E}_{t-1} | \bm{E}_t, \bm{X}_0, \rvc) \propto \bm{E}_t [\bm{Q}_{t}^E]^\top \odot \bm{E}_0 \bar{\bm{Q}}_{t-1}^E$.
Based on this derivation, we only need to create a network $p_{\theta}(\bm{G}_0 | \bm{G}_t, \rvc)$ to predict clean graph $\bm{G}_0$ given noisy data $\bm{G}_t$ and condition $\rvc$.

\subsubsection{External Bond Generation}
\label{sec:bond}

In this stage, we aim to interpolate the distribution $P_{\mathcal{V}_2}$ to $P_\mathcal{B}$ conditioned on $\rvc$, where condition $\rvc$ is $P_\mathcal{X} \times P_\mathcal{G}$.
This is a \textbf{bond-to-bond} generative process, we define $\rvv \sim P_{\mathcal{V}_2}$ as the dummy noisy bond and $\rvb \sim P_\mathcal{B}$ connecting $\rvg$ and $\rvx$, and splice $\rvg$, $\rvx$, and $\rvb$ as a connected graph. We have obtained a trained network $p_{\theta}$ in the last stage, so we freeze $\rvg$ and $\rvx$ in the graph and continue to train $p_{\theta}$.
The principles of the noise-applying and denoising processes in this stage are the same as in Section \colorref{sec:group}, with the only difference being the spaces at both ends of the interpolation.

\subsubsection{Post-adaption}
\label{sec:post-adaption}

\begin{figure}[htbp]
\vspace{-0.15in}
  \centering
  \includegraphics[width=0.8\columnwidth]{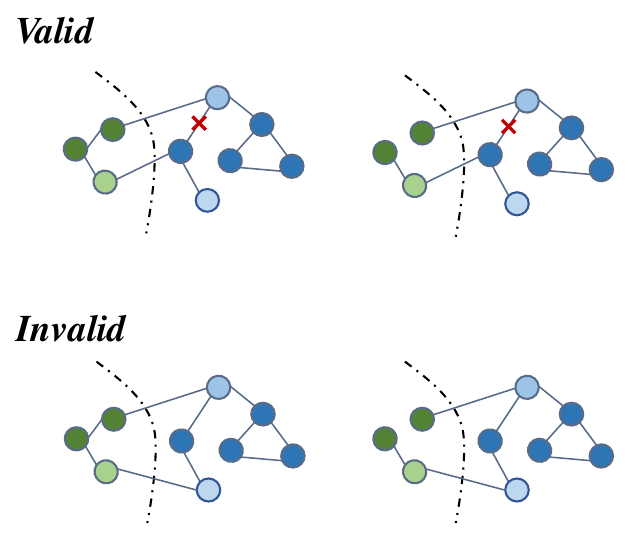}  
  \vspace{-0.1in}
  \caption{Valid and invalid situations of the post-adaption operation. ``\textcolor{red}{x}'' denotes that this bond can be broken manually.}
  \label{img:post}
\vspace{-0.1in}
\end{figure}

Now we get three parts: product (inherent), external group, and external bond. We need to combine these three into final reactants that are chemically legal.

In the traditional template definition of semi-template methods \colorcitep{yan2020retroxpert,shi2020graph,wang2021retroprime}, bonds in the product are first broken to create reaction sites (``reaction center prediction'', usually one bond), and then new leaving groups are generated on them (``synthon completion'').
Back to our steps, we have not yet done any bond-breaking on the product. However, {\bf generated external bonds reveal the reaction site positions on the product, so we can manually break the bond by the legitimacy of reaction sites.} We term this process ``post-adaption'', which serves the same functionality as the ``reaction center prediction'', but with the information of reaction sites, it can be simplified as a rule-based process.

Specifically, traditional semi-template methods obey the following chemical principles:
\vspace{-0.1in}
\begin{itemize}[leftmargin=12px]
    \item (a1) A broken bond is created in the product and the two product atoms corresponding to the broken bond are the reaction site;
    \vspace{-0.05in}
    \item (a2) External groups will be strictly attached to the reaction site.
    \vspace{-0.05in}
\end{itemize}
\vspace{-0.05in}
{\bf Our post-adaptation rule is exactly the inverse process of those chemical principles}:
\vspace{-0.05in}
\begin{itemize}[leftmargin=12px]
    \item (b1) Each generated external bond is attached to one of the atoms of the product. Based on (a2), this product atom must be a reaction site;
    \vspace{-0.05in}
    \item (b2) Based on (a1), a broken bond produces two neighboring reaction sites. Thus, only if two reaction sites are attached to two neighboring product atoms the bond between these two atoms will be broken (the upper part of Figure \colorref{img:post}), otherwise, it is invalid (the lower part of Figure \colorref{img:post}).
    \vspace{-0.05in}
\end{itemize}
Despite being a rule-based process, we find that it can obtain perfect identification performance.
For more complex and extended scenarios, we can also refer to the implementation of ``reaction center prediction'' in previous work and train a predictive model to achieve it. This is the most scalable solution for general scenarios deserving to be explored in future work.

\subsection{Prior Distribution and Interpolation Direction}
\label{sec:prior}

Now we design the task-specific prior distribution (\emph{i.e.} sampling start) $P_{\mathcal{V}_1}$ and $P_{\mathcal{V}_2}$ in the first two stages, and interpolation direction (\emph{i.e.} transitional matrix) $\bm{Q}_t^X$ and $\bm{Q}_t^E$.
We denote $n_g$ and $n_x$ as the atom numbers of the external group $\rvg$ and the product $\rvx$, respectively.
In addition, we cannot predict the exact atom number of external groups in different cases, so we restrict $n_g$ as a constant and create a dummy atom category.
When denoising is complete, the atoms that are still in the dummy category will be deleted, and all remaining atoms constitute the real external group.

\vspace{-0.1in}
\paragraph{Prior Distribution.}
All atoms can start from a single absorbing distribution \colorcitep{austin2021structured} $v_x$ and all bonds can start from $v_e$. 
In the external group generation of stage 1, both atoms and bonds need to be denoised, but in the external bond generation of stage 2, only bonds need to be denoised. Therefore, the two prior distributions can be formulated as
\begin{equation}
    \begin{aligned}
       & P_{\mathcal{V}_1} = p_{v_x}^{|n_g|} \times p_{v_e}^{|n_g|*|n_g|}
       ~~\text{and}~~P_{\mathcal{V}_2} = p_{v_e}^{|n_g|*|n_x|}.
    \end{aligned}
\end{equation}
For all atoms and bonds samples from the dummy state, we set the probability of single distribution as $p_{v_x} = [1,0,0,...,0]^{\top} \in \mathbb{R}^{1 \times (a+1)}$ and $p_{v_e} = [1,0,0,...,0]^{\top} \in \mathbb{R}^{1 \times (b+1)}$, where the first position in the vector denotes the dummy atom (or bond) category and the other positions denote each real categories ($a$ types of atoms and $b$ types of bonds), respectively.

\vspace{-0.1in}
\paragraph{Interpolation Direction.}
For the diffusion model to be reversible, any sample $s = (s_x, s_e) \sim p_\text{data}$ ($p_\text{data}$ denotes the whole data distribution) must converge to a limit distribution $q_{\infty}$ after $t$-step noise-applying, \emph{i,e.}, $q_{\infty} = \lim_{t \rightarrow \infty} s \bar{\bm{Q}}_t$, which in turn is the sampling start.
Therefore, we need to design $\bm{Q}_t^X$ and $\bm{Q}_t^E$ to satisfy that for any atom $s_x$ and bond $s_e$ from the data distribution, $p_{v_x} = \lim_{t \rightarrow \infty} s_x \bar{\bm{Q}}_t^X$ and $p_{v_e} = \lim_{t \rightarrow \infty} s_e \bar{\bm{Q}}_t^E$.
Considering $s_x$ and $s_e$ are one-hot vectors, we compute $\lim_{t \rightarrow \infty} \bar{\bm{Q}}_t^X = \bm{1}_x v_x^{\top}$ and $\lim_{t \rightarrow \infty} \bar{\bm{Q}}_t^E = \bm{1}_e v_e^{\top}$, so a trivial design is
\begin{equation}
    \begin{aligned}
        &\bm{Q}_t^X = \alpha_t \bm{I} + (1 - \alpha_t) \bm{1}_x v_x^{\top},
        ~\bm{Q}_t^E = \alpha_t \bm{I} + (1 - \alpha_t) \bm{1}_e v_e^{\top},
    \end{aligned}
\end{equation}
where $\bm{I}$ is an identity matrix, $\bm{1}_x$ and $\bm{1}_e$ are all-one vectors,
$\alpha_t$ is cosine schedule \colorcitep{nichol2021improved}: 
\begin{equation}
    \bar{\alpha}_t = \cos{(0.5 \pi \frac{t/T+s}{1+s} )^2}.
\end{equation}

\subsection{Denoising Network for Training}
We design $p_{\theta}(\bm{G}_0|\bm{G}_t, \rvc)$ to model $p_{\theta}(\bm{G}_{t-1}|\bm{G}_t, \rvc)$ at the above stages because the latter can be calculated from the former according to Eq.\colorref{eq:parameter}.
At time $t$, we merge the graph $\bm{G}_t = (\bm{X}_t, \bm{E}_t)$ and condition $\rvc$ into a whole graph structure $\bm{G}_w = (\bm{G}_t, \rvc)$, it is treated as the input of $p_{\theta}$. Then we have the output $(p_{\bm{G}'_t}, p_{\rvc'}) = p_{\theta}(\bm{G}_w)$, where $p_{\bm{G}'_t} = (p_{\bm{X}'_t}, p_{\bm{E}'_t})$. 
The training loss of $p_{\theta}(\bm{G}_0|\bm{G}_t, \rvc)$ is:

\vspace{-0.1in}
\begin{small}
\begin{equation}
    \begin{aligned}
        \mathcal{L} = &- \mu \cdot \sum_{x'_t \in \bm{X}'_t, x_0 \in \bm{X}_0} \text{cross-entropy}(p_{x'_t}, x_0) \\ &- \sum_{e'_t \in \bm{E}'_t, e_0 \in \bm{E}_0} \text{cross-entropy}(p_{e'_t}, e_0)
    \end{aligned}
\end{equation}
\end{small}
where $\bm{G}_0 = (\bm{X}_0, \bm{E}_0)$ is the ground truth, and each $x'_t$-$x_0$ / $e'_t$-$e_0$ pair corresponds one-to-one in the graph position.
$\mu$ is a control unit, specifically, in stage 1, $\mu = 0$, and in stage 2, $\mu$ is a hyperparameter to trade off the importance of atoms and bonds. In general, $\mu < 1$.
We use the graph transformer architecture \citep{vignac2022digress,yan2020retroxpert} to design the network. Refer to Appendix \colorref{sec:network} for network details.

\vspace{-0.05in}
\section{Experiments}

\vspace{-0.05in}
\subsection{Setup}

\vspace{-0.05in}
\paragraph{Dataset.}
We conduct experiments on the small-scale USPTO-50K \colorcitep{schneider2016s} and large-scale USPTO-full \colorcitep{usptofull} datasets. The former contains 50K single-step chemical reactions from 10 reaction types, and the latter consists of 760K training data that can demonstrate scalability.
We follow standard splits \colorcitep{schneider2016s} to select 80\%/10\%/10\% of data as training/validation/test sets.

\vspace{-0.12in}
\paragraph{Baseline.}
Our baselines can be divided into three groups:
(i) \textit{Template-based} methods, we choose RetroSim \colorcitep{coley2017computer}, NeuralSym \colorcitep{segler2017neural}, GLN \colorcitep{schneider2016s}, GraphRetro \colorcitep{somnath2020learning}, LocalRetro \colorcitep{chen2021deep}, RetroComposer \colorcitep{yan2022retrocomposer}, and RetroKNN \colorcitep{xie2023retrosynthesis}. 
(ii) \textit{Template-free} methods, we choose Transformer \colorcitep{vaswani2017attention,tetko2020state}, SCROP \colorcitep{zheng2019predicting}, Tied Transformer \colorcitep{kim2021valid}, GTA \colorcitep{seo2021gta}, Graph2SMILES \colorcitep{tu2022permutation}, Chemformer \colorcitep{irwin2022chemformer}, Retroformer \colorcitep{wan2022retroformer}, 
RootAligned \colorcitep{zhong2022root}, RetroDCVAE \colorcitep{he2022modeling}, and RetroBridge \colorcitep{igashov2023retrobridge}.
(iii) \textit{Semi-template} methods, we choose RetroXpert \colorcitep{yan2020retroxpert}, G2G \colorcitep{shi2020graph}, RetroPrime \colorcitep{wang2021retroprime} MEGAN \colorcitep{sacha2021molecule}, and RootAligned \colorcitep{zhong2022root}.

\vspace{-0.12in}
\paragraph{Implementation.}
We use open-source RDKit to construct molecular graphs based on molecular SMILES.
For noise-applying and sampling processes, we set $T_1 = 500$ and $T_2 = 50$. For the training process, we train the graph transformer at 8-card 24G GTX-3090 with a training step of 100K, a batch size of $120$, and an Adam learning rate of $0.0001$, and set $\mu = 0.2$.
In addition, when setting $n_g$, to avoid extreme values that cause sparse distributions during the statistical process, we exclude all samples whose statistic is more than three times the standard deviation from the mean.

\vspace{-0.12in}
\paragraph{Evaluation.}
We follow prior works to adopt top-$k$ {\bf accuracy} as the main evaluation metric. For end-to-end models, beam search is adopted, but it is unfeasible for diffusion models.
Therefore, we set the negative variational lower bound as the ranking score for each generated $\bm{G}_0 = (\bm{X}_0, \bm{E}_0)$:
\vspace{-0.3in}

\begin{equation}
    \begin{aligned}
        \mathrm{Score} = &~\mu \cdot \mathbb{E}_{q(\bm{x}_0)} \mathbb{E}_{q(\bm{x}_t|\bm{x}_0)} [-\log p_{\theta}(x_0|x_t)] \\
        &+ \mathbb{E}_{q(\bm{e}_0)} \mathbb{E}_{q(\bm{e}_t|\bm{e}_0)} [-\log p_{\theta}(e_0|e_t)].
    \end{aligned}
\end{equation}

\vspace{-0.15in}
Note that this evaluation way strictly aligns with the evaluation in \colorcitet{igashov2023retrobridge}.
For each sampling, the smaller the score is, the closer the sample is to the true data distribution. We sample 100 results for each case to rank the scores, and select the $k$ lowest scoring results to compute top-$k$ accuracy.
In addition, we compute top-$k$ {\bf validity} that reflects the legitimacy of the reactants as chemical molecules. It is formulated as $\frac{1}{N \times k} \sum_1^N \sum_1^k \mathrm{isvalid(\bm{G}_0)}$, where $N$ denotes the dataset size.
We also report {\bf round-trip} accuracy and coverage \colorcitep{schwaller2020predicting} for all top-$k$ samples.

\begin{table*}[t]
\centering
\vspace{-0.1in}
\caption{Top-$k$ accuracy for the retrosynthesis task on USPTO-50K and USPTO-full dataset. RetroDiff achieves SOTA among {\it semi-template} methods. For {\it template-free} methods, RetroDiff is competitive, but the validity of generated molecules is significantly higher than theirs (Table \colorref{tab:validity}); For {\it template-based} methods, RetroDiff has a slight performance disadvantage on USPTO-50k, but they perform much poorer than ours on large-scale USPTO-full, exhibiting poor scalability. Note: N/A indicates that the result was not reported in the original paper.}
\label{tab:mainresults}%
\vspace{-0.05in}

\footnotesize
  \renewcommand\arraystretch{1}
  \setlength{\tabcolsep}{3.5mm}{
  \resizebox{2\columnwidth}{!}{
\begin{tabular}{l|cccc|cccc}
\toprule

\multirow{2}{*}{Method} & \multicolumn{4}{c|}{Top-$k$ accuracy in {\bf USPTO-50K}} & \multicolumn{4}{c}{Top-$k$ accuracy in {\bf USPTO-full}} \\
\cmidrule(r){2-9} 

& \multicolumn{1}{c}{$k = 1$}       
& \multicolumn{1}{c}{$k = 3$}       
& \multicolumn{1}{c}{$k = 5$}
& \multicolumn{1}{c|}{$k = 10$}
& \multicolumn{1}{c}{$k = 1$}       
& \multicolumn{1}{c}{$k = 3$}       
& \multicolumn{1}{c}{$k = 5$}
& \multicolumn{1}{c}{$k = 10$}
\\                        

\midrule
\multicolumn{9}{c}{\textbf{Template-based methods}} \\
\midrule
RetroSim \colorcitep{coley2017computer} & 37.3 & 54.7 & 63.3 & 74.1 & 32.8 & - & - & 56.1 \\
NeuralSym \colorcitep{segler2017neural} & 44.4 & 65.3 & 72.4 & 78.9 & 35.8 & - & - & 60.8 \\
GLN \colorcitep{schneider2016s} & 52.5 & 69.0 & 75.6 & 83.7 & 39.0 & 50.1 & 55.3 & 61.3 \\
GraphRetro \colorcitep{somnath2020learning} & 53.7 & 68.3 & 72.2 & 75.5 & \multicolumn{4}{c}{N/A} \\
LocalRetro \colorcitep{chen2021deep} & 53.4 & 77.5 & 85.9 & 92.4 & \multicolumn{4}{c}{N/A} \\
RetroComposer \colorcitep{yan2022retrocomposer} & 54.5 & 77.2 & 83.2 & 87.7 & 41.3 & 53.7 & 56.8 & 63.2 \\
RetroKNN \colorcitep{xie2023retrosynthesis} & 55.3 & 76.9 & 84.3 & 90.8 & \multicolumn{4}{c}{N/A} \\

\midrule
\multicolumn{9}{c}{\textbf{Template-free methods}} \\
\midrule
Transformer \colorcitep{vaswani2017attention} & 42.4 & 58.6 & 63.8 & 67.7 & \multicolumn{4}{c}{N/A} \\
~~~w/ Augmentation ({\it Aug.}) \colorcitep{tetko2020state} & 48.3 & - & 73.4 & 77.4 & 44.4 & - & - & 73.3 \\
SCROP \colorcitep{zheng2019predicting} & 43.7 & 60.0 & 65.2 & 68.7 & \multicolumn{4}{c}{N/A} \\
Tied Transformer \colorcitep{kim2021valid} & 47.1 & 67.1 & 73.1 & 76.3 & \multicolumn{4}{c}{N/A} \\
GTA \colorcitep{seo2021gta} & 51.1 & 67.6 & 74.8 & 81.6 & 46.6 & - & - & 70.4 \\
Graph2SMILES \colorcitep{tu2022permutation} & 52.9 & 66.5 & 70.0 & 72.9 & 45.7 & - & - & 63.4 \\
Chemformer \colorcitep{irwin2022chemformer} & 54.3 & - & 62.3 & 63.0 & \multicolumn{4}{c}{N/A} \\
Retroformer \colorcitep{wan2022retroformer} & 47.9 & 62.9 & 66.6 & 70.7 & \multicolumn{4}{c}{N/A} \\
~~~w/ Augmentation ({\it Aug.}) & 52.9 & 68.2 & 72.5 & 76.4 & \multicolumn{4}{c}{N/A} \\
RootAligned \colorcitep{zhong2022root} & 44.0 & 67.5	& 74.0 & 76.2 & \multicolumn{4}{c}{N/A} \\
~~~w/ 20$\times$ training {\it Aug.}  & 51.5 & 75.0 & 81.0 & 83.0 & \multicolumn{4}{c}{N/A} \\
~~~w/ 20$\times$ training {\it Aug.} + 20$\times$ test {\it Aug.} & 56.0 & 79.1 & 86.1 & 91.0 & \multicolumn{4}{c}{N/A} \\
RetroDCVAE \colorcitep{he2022modeling} & 53.1 & 68.1 & 71.6 & 74.3 & \multicolumn{4}{c}{N/A} \\
RetroBridge \colorcitep{igashov2023retrobridge} & 50.8 & 74.1 & 80.6 & 85.6 & \multicolumn{4}{c}{N/A} \\

\midrule
\multicolumn{9}{c}{\textbf{Semi-template methods}} \\
\midrule
RetroXpert \colorcitep{yan2020retroxpert} & 50.4 & 61.1 & 62.3 & 63.4 & \multicolumn{4}{c}{N/A (Invalid Results\tablefootnote{Data leakage leads to the invalid results. See \url{https://github.com/uta-smile/RetroXpert}.})} \\
G2G \colorcitep{shi2020graph} & 48.9 & 67.6 & 72.5 & 75.5 & \multicolumn{4}{c}{N/A} \\
RetroPrime \colorcitep{wang2021retroprime} & 51.4 & 70.8 & 74.0 & 76.1 & 44.1 & 59.1 & 62.8 & 68.5 \\
MEGAN \colorcitep{sacha2021molecule} & 48.1 & 70.7 & 78.4 & \textbf{86.1} & 33.6 & - & - & 63.9 \\
RootAligned	\colorcitep{zhong2022root} & 49.1 & 68.4 & 75.8 & 82.2 & \multicolumn{4}{c}{N/A} \\
\rowcolor{gray!20}
RetroDiff (ours) & \textbf{52.6} & \textbf{71.2} & \textbf{81.0} & 85.3 & {\bf 46.9} & {\bf 60.4} & {\bf 65.1} &  {\bf 70.3} \\

\bottomrule                   
\end{tabular}}
\vspace{-0.15in}
}

\end{table*}

\vspace{-0.05in}
\subsection{Main Results}
\label{sec:main-results}
\vspace{-0.05in}

We report top-$k$ accuracy and validity in the reactant class unknown setting and compare our method with all strong baselines. Specifically, we categorize our method as a {\it semi-template} method.

\vspace{-0.1in}
\paragraph{Accuracy.}
Table \colorref{tab:mainresults} shows the top-$k$ accuracy results. 

\begin{itemize}[leftmargin=10px]
\vspace{-0.1in}
    \item On the USPTO-50k dataset, our method {\bf outperforms all other competitive semi-template baselines} across different $k$ values, particularly when $k=5$. In addition, our method demonstrates competitive results when compared to the strongest template-free methods. Notably, our method holds a substantial advantage for $k > 1$.
\vspace{-0.05in}
    \item On the USPTO-full dataset, we compare our method with all methods whose original paper reported these results, because we cannot afford the extremely high cost of reproduction. Our method is also {\bf the SOTA of the semi-template methods and outperforms all methods at $k=1,3,5$}. In addition, our top-5 results outperform the top-10 results of most methods. These indicate that our method has a high potential for scalability.
\end{itemize}
\vspace{-0.05in}

In addition, we are surprised to find that the template-based methods exhibit extremely poor accuracies on the large-scale USPTO-full, which are the exact opposite performances on the small-scale USPTO-50k dataset.
This means that {\bf compared with template-based methods, our methods possess greater scalability and are more adaptable to real-world large-scale application scenarios}.

\vspace{-0.05in}
\paragraph{Validity.}
Table \colorref{tab:validity} shows the top-$k$ validity results. We compare our method with some strong template-free baselines whose original paper reported the result.
We don't compare template-based methods because they involve matching existing chemical templates, so they have few validity issues theoretically.

\begin{table}[t]
\vspace{-0.0in}
\begin{center}

\caption{Top-$k$ validity on USPTO-50K dataset of our method and {\it template-free} methods.}
\label{tab:validity}%
\vspace{-0.1in}
\resizebox{1\columnwidth}{!}
{
\begin{tabular}{lcccc}

\toprule
\multirow{2}{*}{Model} & \multicolumn{4}{c}{Top-$k$ validity} \\
\cmidrule(r){2-5} 

& \multicolumn{1}{c}{$k = 1$}       
& \multicolumn{1}{c}{$k = 3$}       
& \multicolumn{1}{c}{$k = 5$}
& \multicolumn{1}{c}{$k = 10$}
\\                        
\cmidrule(r){1-5}
Transformer & 97.2 & 97.9 & 82.4 & 73.1 \\
Graph2SMILES & \textbf{99.4} & 90.9 & 84.9 & 74.9 \\
Retroformer (\textit{Aug.}) & 99.3 & 98.5 & 97.2 & 92.6 \\
\rowcolor{gray!20}
RetroDiff (ours) & 99.2 & \textbf{99.0} & \textbf{97.8} & \textbf{94.3} \\

\bottomrule

\end{tabular}
}

\end{center}
\vspace{-0.25in}
\end{table}

We find that our validity score outperforms all template-free methods, especially as $k$ increases.
As for Retroformer, although it is a template-free method, it integrates the reaction center information defined in semi-template methods during the modeling process. This further reflects that the prior of semi-template methods bring greater validity improvement.
Overall, {\bf compared with template-free methods, our method has a great advantage in generating validity, which reduces unavailability and security risks in practical applications}.

\vspace{-0.1in}
\paragraph{Round-trip Metrics.}

For the top-$k$ samples generated for each product, we assess round-trip accuracy and coverage using the Molecular Transformer model \colorcitep{schwaller2019molecular}.
Round-trip accuracy reflects the percentage of correctly predicted reactants out of all predictions, where a reactant is deemed correct if it matches the ground truth or successfully regenerates the original product.
Round-trip coverage indicates whether at least one correct prediction appears in the top-$k$ samples.
These metrics highlight that a single product can correspond to multiple valid sets of reactants.
Table \colorref{tab:roundtrip} displays the results on the USPTO-50k dataset, showing that RetroDiff maintains strong round-trip performance and achieves state-of-the-art coverage for all $k$ values.

\begin{table}[t]
\centering
\vspace{-0.0in}
\caption{Top-$k$ round-trip results on USPTO-50k.}
\vspace{-0.1in}
\label{tab:roundtrip}%

\footnotesize
  \renewcommand\arraystretch{1}
  \setlength{\tabcolsep}{2mm}{
  \resizebox{1\columnwidth}{!}{
\begin{tabular}{l|ccc|ccc}
\toprule

\multirow{2}{*}{Method} & \multicolumn{3}{c|}{Top-$k$ Coverage} & \multicolumn{3}{c}{Top-$k$ Accuracy}  \\

\cmidrule(r){2-7} 

& \multicolumn{1}{c}{$k = 1$}       
& \multicolumn{1}{c}{$k = 3$}       
& \multicolumn{1}{c|}{$k = 5$}
& \multicolumn{1}{c}{$k = 1$}       
& \multicolumn{1}{c}{$k = 3$}       
& \multicolumn{1}{c}{$k = 5$}

\\                        
\midrule

GLN & 82.5 & 92.0 & 94.0 & 82.5 & 71.0 & 66.2 \\
LocalRetro & 82.1 & 92.3 & 94.7 & 82.1 & 71.0 & 66.7 \\
MEGAN & 78.1 & 88.6 & 91.3 & 78.1 & 67.3 & 61.7 \\
Graph2SMILES & - & - & - & 76.7 & 56.0 & 46.4 \\
Retroformer & - & - & - & 78.6 & 71.8 & 67.1 \\
RetroBridge & 85.1 & 95.7 & 97.1 & {\bf 85.1} & 73.6 & 67.8 \\
\rowcolor{gray!20}
RetroDiff (ours) & {\bf 86.3} & {\bf 96.2} & {\bf 97.6} & 84.5 & {\bf 75.3} & {\bf 69.2} \\

\bottomrule                   
\end{tabular}}
\vspace{-0.25in}
}

\end{table}

\vspace{-0.05in}
\subsection{Ablation}
\vspace{-0.05in}

In this part, We conduct ablation studies to analyze the sub-module performances in each stage, \emph{i.e.}, external group generation and external bond generation.

\vspace{-0.08in}
\paragraph{External Group Generation.}

RetroDiff first generates external groups given raw products.
In traditional {semi-template} methods, the external group generation equates to the synthon completion, commonly addressed in two distinct ways: (i) auto-regressive generation, including encoder-decoder sequence prediction \colorcitep{shi2020graph} and action-state sequence prediction \colorcitep{somnath2020learning}, (ii) finite-space search, where all possible leaving group vocabularies are constructed in a database, followed by maximum likelihood estimation using a classifier \colorcitep{yan2020retroxpert}.
In our setting, the external group generation is treated as non-autoregressive generation.

\begin{table}[t]
\vspace{0.15in}
\begin{center}

\caption{Top-$k$ accuracy of ``external group generation'' sub-module (* indicates the performance of raw synthon completion).}
\label{tab:ablation-group}%
\vspace{-0.1in}

\resizebox{1\columnwidth}{!}
{
\begin{tabular}{lcccc}

\toprule
\multirow{2}{*}{Model} & \multicolumn{4}{c}{Top-$k$ accuracy} \\
\cmidrule(r){2-5} 

& \multicolumn{1}{c}{$k = 1$}       
& \multicolumn{1}{c}{$k = 3$}       
& \multicolumn{1}{c}{$k = 5$}
& \multicolumn{1}{c}{$k = 10$}
\\                        
\cmidrule(r){1-5}

G2G* & 61.1 & 81.5 & 86.7 & 90.0 \\
RetroXpert* & 64.8 & 77.6 & 80.8 & 84.5 \\
\rowcolor{gray!20}
RetroDiff (ours) & 66.5 & 78.4 & 85.0 & 86.4\\

\bottomrule

\end{tabular}
}

\end{center}
\vspace{-0.15in}
\end{table}

Table \colorref{tab:ablation-group} shows the results and we compare the external group generation performance between RetroDiff and the synthon completion performance of other methods.
Our external group generation outperforms the rest of the methods on top-$1$, but not as well as G2G when $k > 1$, albeit within a reasonable margin. A plausible explanation lies in the fact that G2G acquires information about the reaction center when generating the external group, \emph{i.e.}, serial complementation from the reaction sites. In contrast, RetroDiff lacks this specific information, resulting in a slight disadvantage.

\vspace{-0.08in}
\paragraph{External Bond Generation.}

Next, RetroDiff generates external bonds given products and generated external groups.
In the traditional semi-template methods, reaction centers are predicted directly by the classification model, whereas under our template setup, this task equates to a combination of external bond generation and post-adaptation. 
Thus, we conduct a direct comparison between the performance of previous methods in predicting reaction centers and the external bond generation performance of our model.
Table \colorref{tab:ablation-bond} shows the results, indicating that predicting the connecting bond between the product and the external group, and thus deducing the reaction center based on the rule, can achieve higher accuracy than the direct prediction of the reaction center given the product.

\begin{table}[t]
\vspace{-0.in}
\begin{center}

\caption{Top-$k$ accuracy of ``external bond generation'' sub-module (* indicates the performance of raw reaction center prediction).}
\label{tab:ablation-bond}%
\vspace{-0.05in}

\resizebox{1\columnwidth}{!}
{
\begin{tabular}{lcccc}

\toprule
\multirow{2}{*}{Model} & \multicolumn{4}{c}{Top-$k$ accuracy} \\
\cmidrule(r){2-5} 

& \multicolumn{1}{c}{$k = 1$}       
& \multicolumn{1}{c}{$k = 3$}       
& \multicolumn{1}{c}{$k = 5$}
& \multicolumn{1}{c}{$k = 10$}
\\                        
\cmidrule(r){1-5}
G2G* & 61.1 & 81.5 & 86.7 & 90.0 \\
RetroXpert* & 64.3 & - & - & - \\
GraphRetro* & 75.6 & 87.4 & 92.5 & 96.1 \\
\rowcolor{gray!20}
RetroDiff (ours) & 82.3 & 92.4 & 95.5 & 96.8\\

\bottomrule

\end{tabular}
}
\vspace{0.0in}

\end{center}
\vspace{-0.2in}
\end{table}

Specifically, the atom number of the product is denoted as $n$, the bond number as $m$, and the external group atom number as $r$.
Considering a maximum bonding site limit of 4 for an atom (\emph{e.g.} Carbon atom) excluding Hydrogen atoms, we establish the condition $m \leq 2n$.
In the realm of traditional reaction center prediction, the search space size is $m$, whereas, for external bond generation, it is $rn$. Consequently, the complexity of the external bond generation task is higher than that of the reaction center prediction task. However, the external bond generation task leverages molecular information from external groups, expanding the model's ability to search for reaction sites more accurately by incorporating additional chemical insights. Consequently, the observed superior performance of external bond generation over traditional reaction center prediction can be empirically attributed to the enriched chemical information acquired through the former.

\vspace{-0.05in}
\section{Related Work and Discussion}
\vspace{-0.06in}

\paragraph{Related Work.}
Our research focuses on the important biochemical topic of molecular retrosynthesis, which can be categorized into three types: template-based, template-free, and semi-template methods. We summarize the existing retrosynthesis studies in Appendix \colorref{sec:retrosynthesis}.
Additionally, our research methodology involves diffusion models, which have promising applications in the field of molecular generation. Therefore, we introduce existing applications of diffusion models in Appendix \colorref{sec:diffusion}.

\vspace{-0.12in}
\paragraph{Discussion about RetroDiff.}
We also conduct some interesting discussions about RetroDiff in the appendix due to space limits:
(I) Why select the absorbing distribution as the prior (Appendix \colorref{sec:prior-selection});
(II) Why serial multi-stage modeling is used instead of single-stage joint modeling (Appendix \colorref{sec:multi-stage});
(III) Performances on all reaction types (Appendix \colorref{sec:reaction-types});
(IV) Generation-process visualizations (Appendix \colorref{sec:visualization}).

\vspace{-0.06in}
\section{Conclusion}
\vspace{-0.05in}

We introduce RetroDiff, a multi-stage conditional retrosynthesis diffusion model.
Considering maximizing the usage of chemical information in the molecule, we reset the template to decompose the retrosynthesis into external group generation and external bond generation sub-tasks, and set a joint diffusion model to transfer dummy distributions to group and bond distributions serially.
Our method performs the best under the semi-template setting in the accuracy and validity evaluation metrics.
In the future, we will try to extend our RetroDiff to multi-step retrosynthesis scenarios.

\section*{Acknowledgements}
This work is supported by the the Natural Science Foundation of China (Grant No. 62376133), Beijing Nova Program (20240484682) and Wuxi Research Institute of Applied Technologies, Tsinghua University under Grant 20242001120.

\bibliographystyle{Ref}
\bibliography{Ref}

\section*{Checklist}

 \begin{enumerate}

 \item For all models and algorithms presented, check if you include:
 \begin{enumerate}
   \item A clear description of the mathematical setting, assumptions, algorithm, and/or model. \textcolor{blue}{[Yes]}
   \item An analysis of the properties and complexity (time, space, sample size) of any algorithm. \textcolor{blue}{[Not Applicable]}
   \item (Optional) Anonymized source code, with specification of all dependencies, including external libraries.
 \end{enumerate}

 \item For any theoretical claim, check if you include:
 \begin{enumerate}
   \item Statements of the full set of assumptions of all theoretical results. \textcolor{blue}{[Not Applicable]}
   \item Complete proofs of all theoretical results. \textcolor{blue}{[Not Applicable]}
   \item Clear explanations of any assumptions. \textcolor{blue}{[Not Applicable]}
 \end{enumerate}

 \item For all figures and tables that present empirical results, check if you include:
 \begin{enumerate}
   \item The code, data, and instructions needed to reproduce the main experimental results (either in the supplemental material or as a URL). \textcolor{blue}{[Yes]}
   \item All the training details (e.g., data splits, hyperparameters, how they were chosen). \textcolor{blue}{[Yes]}
         \item A clear definition of the specific measure or statistics and error bars (e.g., with respect to the random seed after running experiments multiple times). \textcolor{blue}{[Not Applicable]}
         \item A description of the computing infrastructure used. (e.g., type of GPUs, internal cluster, or cloud provider). \textcolor{blue}{[Yes]}
 \end{enumerate}

 \item If you are using existing assets (e.g., code, data, models) or curating/releasing new assets, check if you include:
 \begin{enumerate}
   \item Citations of the creator If your work uses existing assets. \textcolor{blue}{[Yes]}
   \item The license information of the assets, if applicable. \textcolor{blue}{[Not Applicable]}
   \item New assets either in the supplemental material or as a URL, if applicable. \textcolor{blue}{[Not Applicable]}
   \item Information about consent from data providers/curators. \textcolor{blue}{[Yes]}
   \item Discussion of sensible content if applicable, e.g., personally identifiable information or offensive content. \textcolor{blue}{[Not Applicable]}
 \end{enumerate}

 \item If you used crowdsourcing or conducted research with human subjects, check if you include:
 \begin{enumerate}
   \item The full text of instructions given to participants and screenshots. \textcolor{blue}{[Not Applicable]}
   \item Descriptions of potential participant risks, with links to Institutional Review Board (IRB) approvals if applicable. \textcolor{blue}{[Not Applicable]}
   \item The estimated hourly wage paid to participants and the total amount spent on participant compensation. \textcolor{blue}{[Not Applicable]}
 \end{enumerate}

 \end{enumerate}

\onecolumn
\aistatstitle{RetroDiff: Retrosynthesis as Multi-stage Distribution Interpolation \\
Supplementary Materials}

\appendix

\section{Related Work}

\subsection{Retrosynthesis Prediction}
\label{sec:retrosynthesis}
Existing methods of retrosynthesis prediction can be broadly categorized into three groups: 
(i) \textit{Template-based} methods retrieve the best match reaction template for a target molecule from a large-scale chemical database, they focus on computing the similarity scores between target molecules and templates using either plain rules \colorcitep{coley2017computer} or neural networks \colorcitep{schneider2016s,somnath2020learning,chen2021deep}.
(ii) \textit{Template-free} methods adopt end-to-end generative models to directly obtain final reactants given products \colorcitep{zheng2019predicting,kim2021valid,seo2021gta,tu2022permutation,wan2022retroformer}. Despite the efficiencies of data-driven methods, the chemical prior has been ignored.
(iii) \textit{Semi-template} methods combine the advantages of the above two approaches, they split the task into two parts, \emph{i.e.}, reaction center prediction and synthon correction \colorcitep{yan2020retroxpert,shi2020graph,wang2021retroprime}, followed by serial modeling using a classification model and a generative model, respectively.

\subsection{Diffusion Models in Molecules}
\label{sec:diffusion}
Diffusion models \colorcitep{sohl2015deep,ho2020denoising} is a class of score-based generative models \colorcitep{song2019generative}, whose goal is to learn the latent structure of a dataset by modeling how data points diffuse through the latent space.
Since the generalized discrete diffusion model \colorcitep{austin2021structured} and the discrete graph diffusion model \colorcitep{vignac2022digress} have been proposed, the molecular design field began to use them extensively, such as molecular conformation \colorcitep{xu2021geodiff}, molecular docking \colorcitep{corso2022diffdock}, and molecular linking \colorcitep{igashov2022equivariant}. 
To our knowledge, we are the first to apply discrete diffusion models to the retrosynthesis prediction task.
\colorcitep{igashov2023retrobridge} have done similar work using a diffusion model, and they achieve the template-free retrosynthesis prediction. However, our method performs better and has stronger chemical interpretability.
\vspace{-0.05in}

\section{Denoising Network for Training}
\label{sec:network}

\subsection{Network Architecture}
\label{sec:architecture}

\begin{figure*}[htbp]
  \centering
  \includegraphics[width=0.8\columnwidth]{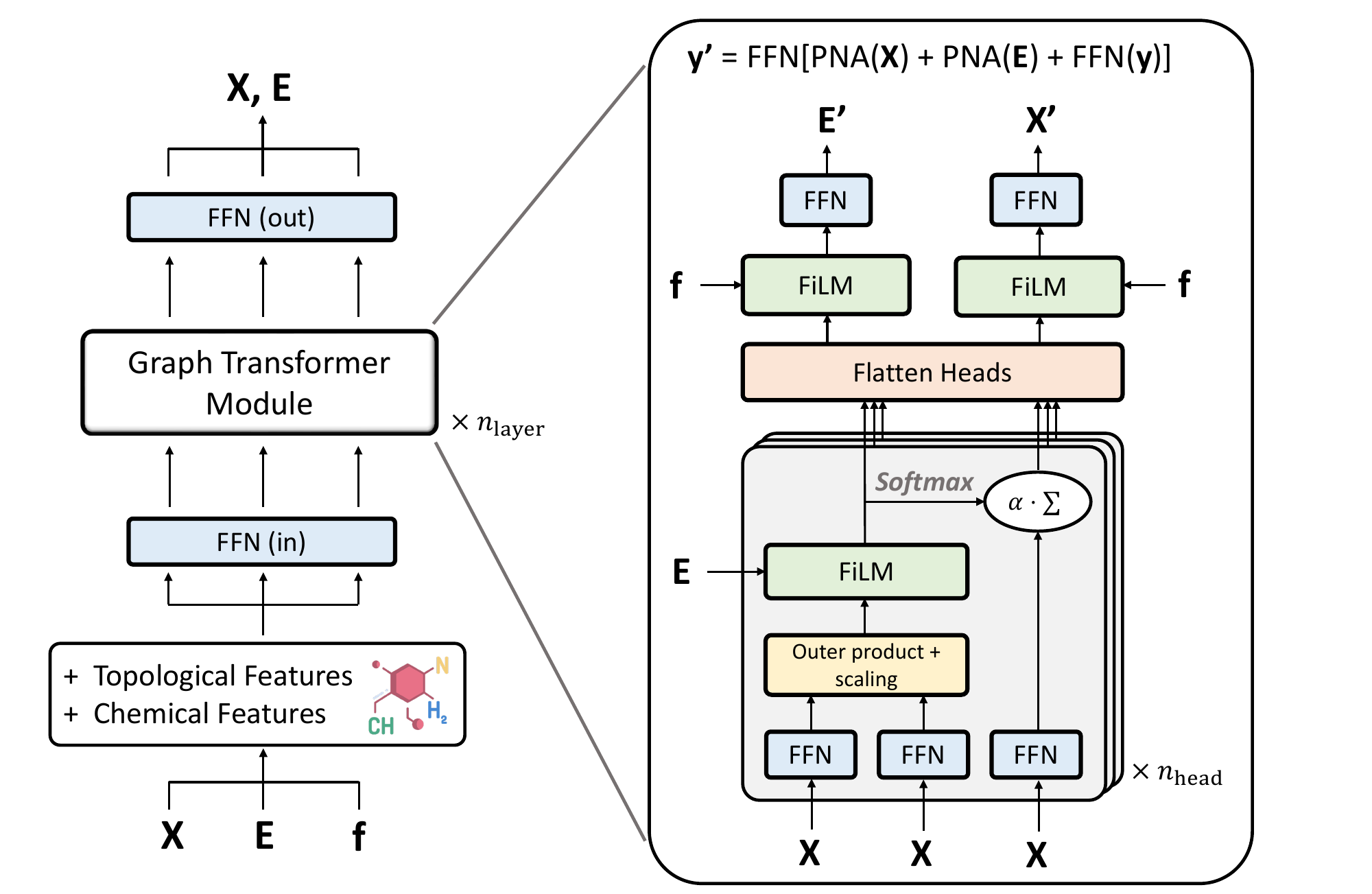}  
  \vspace{-0.0in}
  \caption{The whole architecture (left) of the denoising network for training with graph transformer modules (right). $\bm{X}, \bm{E}, \bm{y}$ denote the atom features, bond features, and global features, respectively. $\mathrm{FiLM}(\bm{M}_1, \bm{M}_2) = \bm{M}_1 \bm{W}_1 + (\bm{M}_2 \bm{W}_2) \odot \bm{M}_2 + \bm{M}_2$, where $\bm{W}_1, \bm{W}_2$ are learnable. $\mathrm{PNA}(\bm{M}) = [\mathrm{max}(\bm{M}) \circ \mathrm{min}(\bm{M}) \circ \mathrm{mean}(\bm{M}) \circ \mathrm{std}(\bm{M})] \bm{W}$, where $\bm{W}$ is learnable.}
  \label{img:network}
\end{figure*}

The overall architecture and the graph transformer module for each layer are shown in Figure \colorref{img:network}.
Specifically, we add the global feature $\bm{f}$ to the input, so the final input at time $t$ is $(\bm{G}_w, \bm{f}) = (\bm{X}, \bm{E}, \bm{f})$.
For the global feature $\bm{f}$, we obtain the topological features and chemical features (Details can be seen in Appendix \colorref{sec:feature}) of this molecular graph to splice with the original features.
After the pre-processing, $\bm{G} = (\bm{X}, \bm{E}, \bm{f})$ is input to a feed-forward network to be encoded, then it will pass serially through the $n_{\mathrm{layer}}$ graph transformer modules. Finally, another feed-forward network is set to decode the graph features, the output is the final prediction result.

\subsection{Additional Input Features}
\label{sec:feature}
To fully explore the potential features of a molecular graph, we can analyze it from two perspectives: topological features and chemical features following \colorcitep{vignac2022digress,yan2020retroxpert}.

\paragraph{Topological Features.}

We focus on two useful topological features.
First is the \textbf{spectral features}, we first compute some graph-level features that relate to the eigenvalues of the graph Laplacian: the number of connected components (given by the multiplicity of eigenvalue 0), as well as the 5 first nonzero eigenvalues. We then add node-level features relative to the graph eigenvectors: an estimation of the biggest connected component (using the eigenvectors associated with eigenvalue 0), as well as the two first eigenvectors associated with non-zero eigenvalues.

Second is the \textbf{cycle detection}. To further refine it, we split it into node-level and graph-level features.
For node-level features, we compute how many $k$-cycles this node belongs to, where $3 \leq k \leq 5$. The feature formulas are as follows:
\begin{equation}
    \begin{aligned}
        & \bm{X}_3 = \mathrm{diag}(\bm{A}^3) / 2, \\
        & \bm{X}_4 = (\mathrm{diag}(\bm{A}^4) - \bm{d}(\bm{d} - 1) - \bm{A}(\bm{d}\bm{1}_n^{\top}) \bm{1}_n) / 2, \\
        & \bm{X}_5 = (\mathrm{diag}(\bm{A}^5) - 2 \mathrm{diag}(\bm{A}^3) \odot \bm{d} - \bm{A} (\mathrm{diag}(\bm{A}^3) \bm{1}_n^{\top})\bm{1}_n + \mathrm{diag}(\bm{A}^3)) / 2, \\
    \end{aligned}
\end{equation}
where $\bm{d}$ denotes the vector containing node degrees.
For graph-level features, we compute how many $k$-cycles this graph contains, where $3 \leq k \leq 6$. The feature formulas are as follows:
\begin{equation}
    \begin{aligned}
        & \bm{y}_3 = \bm{X}_3^{\top} \bm{1}_n / 3, \\
        & \bm{y}_4 = \bm{X}_4^{\top} \bm{1}_n / 4, \\
        & \bm{y}_5 = \bm{X}_5^{\top} \bm{1}_n / 5, \\
        & \bm{y}_6 = \mathrm{Tr}(\bm{A}^6) - 3 \mathrm{Tr}(\bm{A}^3 \odot \bm{A}^3) + 9 ||\bm{A} (\bm{A}^2 \odot \bm{A}^2)||_F - 6 \mathrm{diag}(\bm{A}^2)^{\top} \mathrm{diag}(\bm{A}^4) \\ & \qquad + 6 \mathrm{Tr}(\bm{A}^4) - 4 \mathrm{Tr}(\bm{A}^3) + 4 \mathrm{Tr}(\bm{A}^2 \dot{\bm{A}}^2 \odot \bm{A}^2) + 3 ||\bm{A}^3||_F - 12 (\bm{A}^2 \odot \bm{A}^2) + 4 \mathrm{Tr}(\bm{A}^2), \\
    \end{aligned}
\end{equation}
where $||\cdot||_F$ is Frobenius norm.

\paragraph{Chemical Features.}

There are two useful chemical features. 
First is the \textbf{atom valency}, which can be concatenated to the atom features $\bm{X}$. 
Second is the \textbf{molecular weight}, which can be concatenated to the global features $\bm{y}$.

\section{Discussion: Extended Analysis}

\subsection{Prior Distribution Selection}
\label{sec:prior-selection}

In our setting of the prior distribution, we choose the absorbing distribution \colorcitep{austin2021structured}, which is a special kind of marginal distribution \colorcitep{vignac2022digress} from a generalized perspective, \emph{i.e.} collapses the probabilities at all positions to those at one of them, making the initial state practically deterministic.
We use it because the noisy graph we define is a dummy graph, which means that the probability of the empty category is 1 and others 0.

Another common prior distribution is the uniform distribution, meaning the sampling starts from a completely random state. We compare the two distributions in the whole process, the external group generation process, and the external bond generation process in Table \colorref{tab:prior-distribution}.

\begin{table}[htbp]
\begin{center}
\caption{Top-$k$ accuracy under different prior distributions.}
\label{tab:prior-distribution}
\resizebox{0.98\columnwidth}{!}
{
\begin{tabular}{lcccc|cccc|cccc}

\toprule
\multirow{2}{*}{Prior Dist.}& \multicolumn{12}{c}{Top-$k$ accuracy} 
\\
\cmidrule(r){2-13} 

& \multicolumn{1}{c}{$k = 1$}       
& \multicolumn{1}{c}{$k = 3$}       
& \multicolumn{1}{c}{$k = 5$}
& \multicolumn{1}{c|}{$k = 10$}
& \multicolumn{1}{c}{$k = 1$}       
& \multicolumn{1}{c}{$k = 3$}       
& \multicolumn{1}{c}{$k = 5$}
& \multicolumn{1}{c|}{$k = 10$}
& \multicolumn{1}{c}{$k = 1$}       
& \multicolumn{1}{c}{$k = 3$}       
& \multicolumn{1}{c}{$k = 5$}
& \multicolumn{1}{c}{$k = 10$}
\\        

\cmidrule(r){2-13} 
& \multicolumn{4}{c|}{Overall} & \multicolumn{4}{c|}{External Group Generation} & \multicolumn{4}{c}{External Bond Generation} \\

\midrule
Uniform & 51.7 & 70.1 & 79.6 & 83.2 & 66.2 & {\bf 78.8} & 84.6 & 86.2 & 80.2 & 89.8 & 91.7 & 92.9 \\
Absorbing & {\bf 52.6} & {\bf 71.2} & {\bf 81.0} & {\bf 85.3} & {\bf 66.5} & 78.4 & {\bf 85.0} & {\bf 86.4} & {\bf 82.3} & {\bf 92.4} & {\bf 95.5} & {\bf 96.8} \\

\bottomrule

\end{tabular}
}

\end{center}
\vspace{-0.0in}
\end{table}

We find that absorbing distribution performs slightly better than uniform distribution, and the difference mainly appears in the external bond generation. Since most of the external bonds are EMPTY in the ground truth, which means that the bond categories of most positions do not change during the diffusion process, setting an absolute distribution distribution helps faster convergence and accurate learning. 
From this perspective, we find that a low-entropy marginal prior distribution (\emph{e.g.} absorbing distribution) is more suitable for predictive tasks like external bond generation.

\subsection{Single-stage Diffusion {\it vs.} Multi-stage Diffusion}
\label{sec:multi-stage}

There are three modes of the modeling approach: (1) joint modeling, where the group and the bond are generated at the same time; (2) generating the group first and then the bond; (3) generating the bond first and then the group.
The first is the single-stage diffusion, and the latter two are the multi-stage diffusions.
We compare the performances of the three modes, as shown in Table \colorref{tab:mode}.

\begin{table}[htbp]
\begin{center}
\caption{Top-$k$ accuracy under different diffusion modeling modes.}
\resizebox{0.65\columnwidth}{!}
{
\begin{tabular}{lcccc}

\toprule
\multirow{2}{*}{Modeling Mode} & \multicolumn{4}{c}{Top-$k$ accuracy} \\
\cmidrule(r){2-5} 

& \multicolumn{1}{c}{$k = 1$}       
& \multicolumn{1}{c}{$k = 3$}       
& \multicolumn{1}{c}{$k = 5$}
& \multicolumn{1}{c}{$k = 10$}
\\                        
\midrule

Joint Modeling & 51.3 & 69.6 & 79.8 & 84.1\\
First Group, Then Bond & {\bf 52.6} & {\bf 71.2} & {\bf 81.0} & {\bf 85.3} \\
First Bond, Then Group & 49.8 & 66.1 & 76.7 & 81.4 \\

\bottomrule

\end{tabular}
}

\label{tab:mode}%
\end{center}
\end{table}

Previous work \colorcitep{jo2022score} concluded that joint modeling has better performance than ``marginal then conditional'' serial modeling. However, in our initial explorations, we find that mode (2), the serial mode of generating groups first and then bonds, was more effective for the retrosynthesis task.
We speculate that the underlying reason is that {\bf the information discrepancy of groups and bonds is large}. The information of additional bond information might not affect the group generation too much, whereas additional group information might largely determine the formation of a bond. Thus, generating groups first brings a large positive effect on bond generation.
Therefore, we adopt the multi-stage diffusion in our work.
This is an interesting phenomenon whose underlying reasons deserve to be explored in future work.

\subsection{Performance in All Reaction Types}
\label{sec:reaction-types}

We list all reaction classes on the USPTO-50k dataset, and report the top-$k$ accuracy of each reaction class when trained with reaction class unknown in Table \colorref{tab:allreaction}.
From the results, we have the following analyses:
\begin{itemize}[leftmargin=20px]
    \item For some reactions, such as functional group addition, oxidations, and the protections, the accuracy is significantly higher than the average.
    \item For some other reactions, such as functional group interconversion and C-C bond formation, the accuracy is significantly lower than the average.
    \item These observation helps us better understand RetroDiff's strengths and limitations on different reactions, improving the interpretability of RetroDiff.
\end{itemize}

\begin{table*}[t]
\centering
\caption{Top-$k$ accuracy on all reaction classes.}
\label{tab:allreaction}%

\footnotesize
  \renewcommand\arraystretch{1}
  \setlength{\tabcolsep}{3mm}{
  \resizebox{1\columnwidth}{!}{
\begin{tabular}{ll|cccc}
\toprule

\multirow{2}{*}{Reaction Class} & \multirow{2}{*}{Reaction Fraction(\%)} & \multicolumn{4}{c}{Top-$k$ accuracy in {\bf USPTO-50K}} \\

\cmidrule(r){3-6} 

& & \multicolumn{1}{c}{$k = 1$}       
& \multicolumn{1}{c}{$k = 3$}       
& \multicolumn{1}{c}{$k = 5$}
& \multicolumn{1}{c}{$k = 10$}

\\                        
\midrule

heteroatom alkylation and arylation	& 30.3 & 51.4 & 68.4 & 80.0 & 84.6 \\
acylation and related processes & 23.8 & 60.2 & 78.0 & 87.6 & 90.2 \\
deprotections & 16.5 & 49.1 & 75.7 & 82.4 & 88.6 \\
C-C bond formation & 11.3 & 41.7 & 60.3 & 71.2 & 75.3 \\
reductions & 9.2 & 58.9 & 76.8 & 82.5 & 88.2 \\
functional group interconversion & 3.7 & 30.5 & 51.0 & 62.7 & 68.4 \\
heterocycle formation & 1.8 & 47.2 & 68.3 & 76.4 & 78.2 \\
oxidation & 1.6 & 73.6 & 83.4 & 90.8 & 91.7 \\
protections & 1.3 & 72.1 & 87.8 & 88.2 & 89.8 \\
functional group addition & 0.5 & 84.0 & 84.0 & 86.3 & 88.0 \\

\bottomrule                   
\end{tabular}}
\vspace{-0.05in}
}

\end{table*}

\subsection{Case Study via Visualization}
\label{sec:visualization}

\begin{figure*}[htb]
  \centering
  \includegraphics[width=1\textwidth]{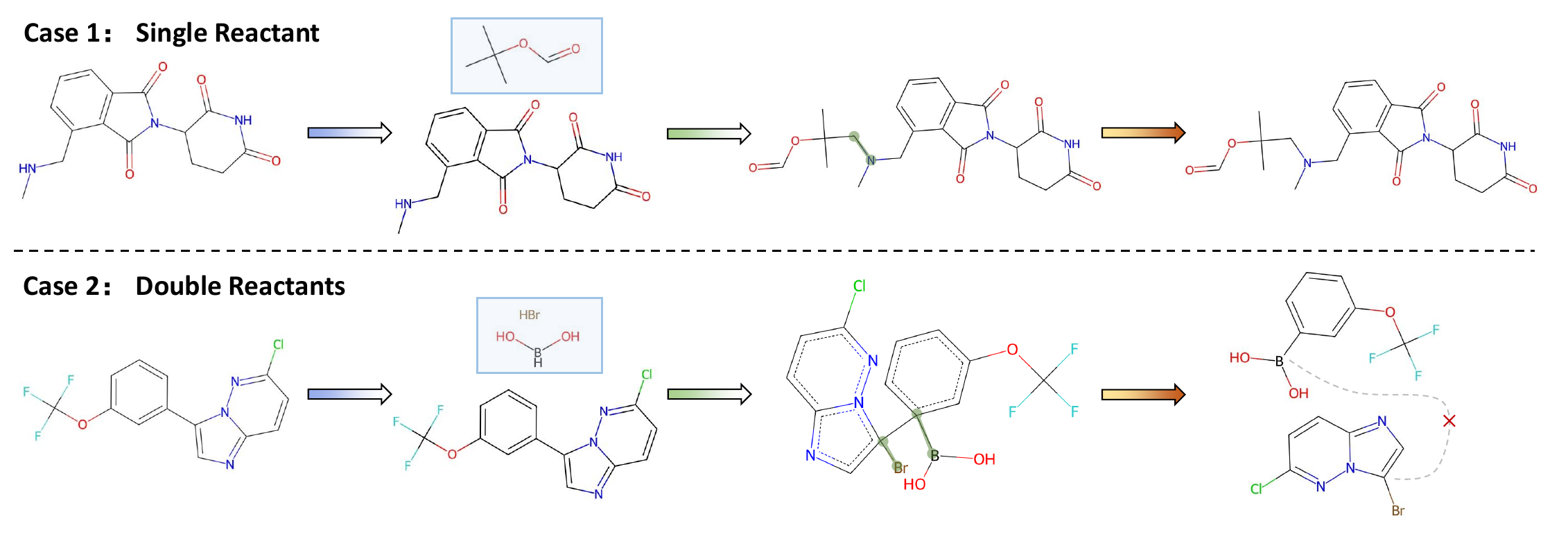}  
  \vspace{-0.2in}
  \caption{Successful cases produced by RetroDiff on the retrosynthesis task.}
  \label{img:successful-cases}
\end{figure*}

\begin{figure*}[htb]
  \centering
  \includegraphics[width=1\textwidth]{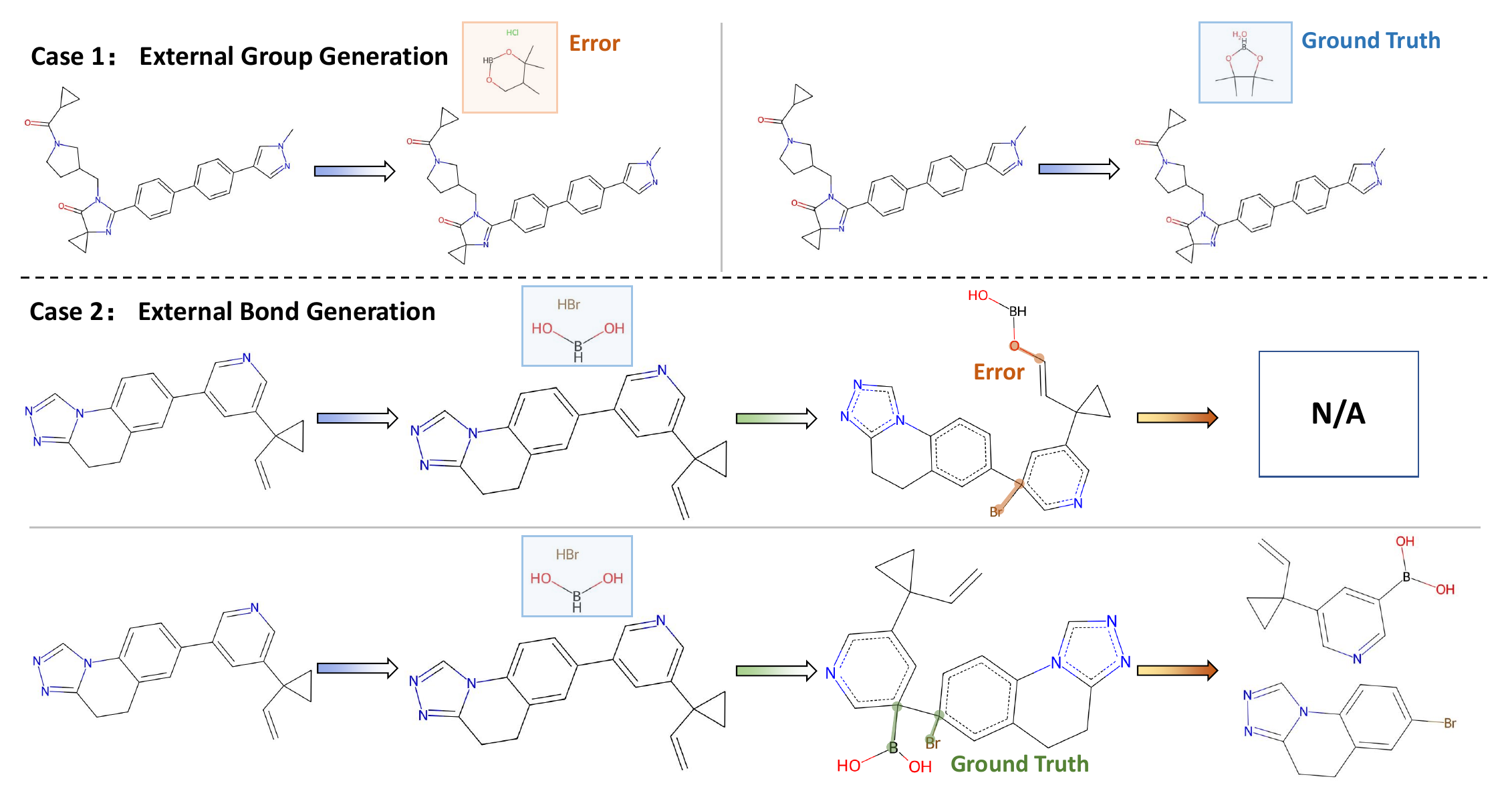}  
  \vspace{-0.2in}
  \caption{Failed cases produced by RetroDiff on the retrosynthesis task.}
  \label{img:failed-cases}
\end{figure*}

In this part, we present visualizations of both successful and failed cases to provide an intuitive analysis of RetroDiff's mechanisms.
Figure \colorref{img:successful-cases} illustrates instances of success, featuring external groups delineated by blue shaded boxes and external bonds highlighted in green.
Conversely, Figure \colorref{img:failed-cases} showcases failed cases, revealing two prevalent situations associated with higher error rates: (i) elevated error rates are observed when the external group size is substantial, leading to biases in the prediction of bonds between atoms, and (ii) for external bond predictions, inaccuracies in predicting reaction sites on the product contribute to ineffective post-adaptation of reaction centers.

\end{document}